\documentclass[10pt,twocolumn,letterpaper]{article}
       
\usepackage[pagenumbers]{cvpr} 

\definecolor{cvprblue}{rgb}{0.21,0.49,0.74}
\usepackage[pagebackref,breaklinks,colorlinks,allcolors=cvprblue]{hyperref}
\usepackage[acronym, shortcuts]{glossaries-extra}

\glssetcategoryattribute{acronym}{nohyper}{true}
\setabbreviationstyle[acronym]{long-short}

\newacronym{llm}{LLM}{Large Language Model}
\newacronym{eqa}{EQA}{Embodied Question Answering}
\newacronym{vqa}{VQA}{Visual Question Answering}
\newacronym{vlfm}{VLFM}{Vision-Language Frontier Maps}
\newacronym{sota}{SoA}{state-of-the-art}

\usepackage{graphicx}
\usepackage{booktabs}
\usepackage{amssymb}
\usepackage{pifont}
\usepackage{array,multirow}
\usepackage{colortbl}
\usepackage{wrapfig}
\usepackage{comment}

\newcommand{\approach}{TANGO}
\def\paperID{} 
\def\confName{CVPR}
\def\confYear{2025}

\title{TANGO: Training-free Embodied AI Agents for Open-world Tasks}

\author{Filippo Ziliotto$^{1,2}$ \qquad Tommaso Campari$^{2}$ \qquad Luciano Serafini$^2$ \qquad Lamberto Ballan$^1$ \\
\\
$^{1}$ University of Padova \\ $^{2}$ Fondazione Bruno Kessler (FBK) \\}

\begin{document}

\maketitle

\begin{abstract}
Large Language Models (LLMs) have demonstrated excellent capabilities in composing various modules together to create programs that can perform complex reasoning tasks on images. In this paper, we propose TANGO, an approach that extends the program composition via LLMs already observed for images, aiming to integrate those capabilities into embodied agents capable of observing and acting in the world. Specifically, by employing a simple PointGoal Navigation model combined with a memory-based exploration policy as a foundational primitive for guiding an agent through the world, we show how a single model can address diverse tasks without additional training.
We task an LLM with composing the provided primitives to solve a specific task, using only a few in-context examples in the prompt.
We evaluate our approach on three key Embodied AI tasks: Open-Set ObjectGoal Navigation, Multi-Modal Lifelong Navigation, and Open Embodied Question Answering, achieving state-of-the-art results without any specific fine-tuning in challenging zero-shot scenarios.  
\end{abstract}


\section{Introduction}
\label{sec:intro}
\begin{figure}[ht]
\begin{center}
\includegraphics[width=.46\textwidth]{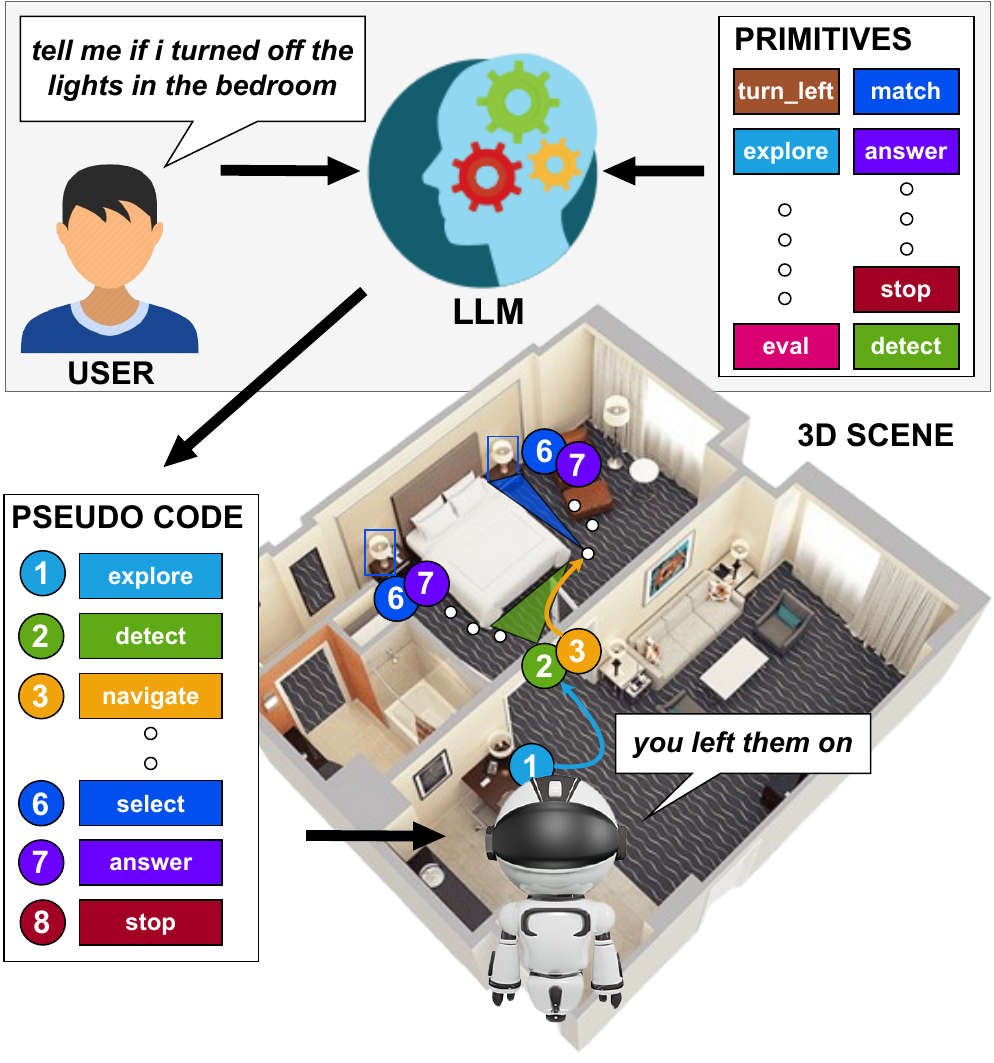}
\caption{We introduce \textbf{\approach}, a modular neuro-symbolic system for compositional embodied visual navigation. Given a few examples of natural language instructions and the corresponding programs composed of action primitives, \approach\ can generate executable programs, enabling the agent to perform multiple tasks within a 3D environment.}
\label{img:teaser}
\vspace{-15pt}
\end{center}
\end{figure}

\acp{llm} have gained significant attention in the field of AI due to their remarkable capability to generalize across unseen tasks \cite{gpt, team2023gemini, jiang2023mistral, touvron2023llama}. 
Systems like VisProg and ViperGPT~\cite{visprog,vipergpt} demonstrated strong performance on a broad range of vision-and-language tasks by just relying on a few numbers of in-context examples to generate complex compositional programs without requiring any specific training. 
Nevertheless, the evaluation of these concepts to Embodied AI, particularly for navigation tasks, remains largely unexplored. This approach may be crucial in building agents capable to navigate and operate efficiently in unfamiliar environments.

Previously, methods such as Neural Module Networks (NMN)~\cite{nmn2016,hu2017learning} have demonstrated promising compositional properties for high-level tasks like visual question answering, via end-to-end training of networks and specialized, differentiable neural modules. However, they required the combination of semantic parsers and predefined templates to learn to solve the compositional tasks. 
Alternative approaches leverage \acp{llm} through predefined task modules or APIs for action execution~\cite{zhou2024navgpt, wu2023embodied, huang2022language, huang2022inner,liang2023code}. While effective, these methods often lack a compositional structure to address multiple tasks or primarily focus on manipulation and robotic planning, with limited emphasis on navigation.

In this work, we present \approach\ (see Figure~\ref{img:teaser}) a novel neuro-symbolic compositional approach which utilizes primitives and employs a \ac{llm} as a planner to sequence these primitives within photorealistic 3D environments, where agents must perceive and act. This framework integrates high-level planning with low-level action execution, without the need for training. \approach\ makes use of diverse modules designed for visual navigation and question-answering tasks, resulting in a system that seamlessly addresses both challenges while achieving SoA performance without requiring any prior adjustments.
By providing a few in-context examples that show how to tackle multiple tasks, \approach\ is capable of generalizing to the specific task at hand. This is facilitated by the \ac{llm}, which effectively combines the individual modules available within the \approach\ system.
Moreover, it extends \cite{yokoyama2024vlfm} exploration policy by incorporating a memory mechanism in the form of a stored feature map that retains information about previously explored areas, supporting for efficient life-long navigation tasks.

To demonstrate the flexibility of the proposed framework, \approach\ has been tested on three popular benchmarks: namely, \emph{i}) Open-Vocabulary ObjectGoal Navigation~\cite{yokoyama2024hm3d},  \emph{ii}) Multi-Modal Lifelong Navigation~\cite{khanna2024goat}, and \emph{iii}) Embodied Question Answering \cite{majumdar2024openeqa, das2018embodied}. Results match or surpass previous approaches, without requiring specialized training.
Summing up, the contributions of this paper are as follows:
\begin{itemize} 
\item  Introduction of a neuro-symbolic compositional LLM-based framework for EAI leveraging specialized primitive modules.
\item Demonstration of robust generalization capabilities across multiple tasks without the need for any specific training or fine-tuning. 
\item Extension of the exploration policy presented in \cite{yokoyama2024vlfm} to multi-goal scenarios through the incorporation of a memory mechanism stored as a feature vector map.
\item Achievement of state-of-the-art results, underscoring the effectiveness of the proposed framework. 
\end{itemize}



\section{Related Works}
\label{sec:relworks}
Our work takes inspiration from modular approaches for visual tasks, like the seminal Neural Module Networks~\cite{nmn2016, hu2017learning}, as well as the recent VisProg~\cite{visprog} and ViperGPT~\cite{vipergpt} frameworks. The main contribution of this paper is to extend these ideas to embodied visual navigation, that usually require heavy end-to-end learning, although modular approaches have recently shown promising results~\cite{modlearn, campari2022online}.
Therefore, we discuss prior work in the area of embodied AI, the use of language models in robotics, and program generation approaches for image recognition tasks.

\paragraph{Embodied AI and Visual Navigation.}
The field of Embodied AI has recently undergone a paradigm shift, fuelled by the emergence of highly efficient simulators~\cite{savva2017minos, savva2019habitat, kolve2017ai2, shen2021igibson, puig2023habitat}. These simulators enable processing numerous parallel simulations in photorealistic indoor environments, facilitating large-scale testing that was otherwise challenging in classical robotics. 
Alongside these simulators, a large variety of tasks (and benchmarks) has been proposed: PointGoal Navigation~\cite{wijmans2019dd}, where an agent navigates from point A to point B in an unknown environment; ObjectGoal Navigation~\cite{batra2020objectnav}, requiring the agent to locate and navigate to an object in the scene; Instance-Image Goal Navigation~\cite{krantz2022instance}, akin to ObjectGoal Navigation, but in which the agent should find a specific object depicted in a given image; Multi-Modal Lifelong Navigation~\cite{khanna2024goat} involving navigating to a sequence of target objects that can be specified through labels, images, or textual descriptions; Embodied Question Answering~\cite{majumdar2024openeqa, das2018embodied}, in which an agent navigates an environment to gather information needed to answer a question, and Vision and Language Navigation (VLN)~\cite{anderson2018vision}.  

Various approaches have been proposed to tackle these tasks, primarily falling into three categories: end-to-end~\cite{ye2020auxiliary, ye2021auxiliary, campari2020exploiting}, modular~\cite{chaplot2020learning, chaplot2020neural, chaplot2020object, ramakrishnan2022poni, krantz2023navigating, raychaudhuri2024mopa} and LLM-based~\cite{zhou2024navgpt,dorbala2023can,zhou2025navgpt, yul3mvn, li2024tina}. 
However, the main limitation of these paradigms is their dependence on task-specific architectures. 
In the case of end-to-end approaches, a model is trained for a specific problem, often requiring days of training, such as in \cite{wijmans2019dd}, where a policy for PointGoal Navigation was trained for 2.5 billion steps. 
To adapt the learned model to a new problem, adjustments and retraining are necessary, and this comes with a high cost. 
Modular approaches share a similar challenge, usually relying on policies designed for specific tasks, although common elements, such as exploration and navigation modules, are often shared~\cite{chaplot2020object, chaplot2020learning, ramakrishnan2022poni, krantz2023navigating, raychaudhuri2024mopa}. Adapting these solutions to different tasks requires a manual adjustment of the approach to fit the new domain, often involving the addition or modification of modules. 

\paragraph{Language models for Robotics}
In robotics, foundation models (\acp{llm}) trained on vast internet-scale datasets~\cite{gpt, team2023gemini, jiang2023mistral, touvron2023llama} have the potential to equip robots with real-world priors and advanced reasoning abilities without the need for extensive task-specific training. Early approaches equipped agents with learned language embeddings, requiring large amounts of training data~\cite{bing2023meta,liang2023code}. Recent studies, on the other hand, have explored zero-shot and few-shot solutions mainly focusing on robotic planning and manipulation tasks~\cite{huang2022language, liang2023code, huang2022inner}. 
In the context of visual navigation, LLM-based approaches leverage the powerful \acp{llm} priors and reasoning capabilities to guide navigation within the environment~\cite{zhou2024navgpt,zhou2025navgpt,dorbala2023can, yul3mvn, li2024tina}; however, despite their remarkable contributions, they often lack a modular design that allows for flexible integration and extensibility, as they are typically optimized for single tasks rather than a comprehensive, adaptable framework. 

In contrast, our work deviates from these paradigms by relying on a series of diverse pre-trained modules that, thanks to a \ac{llm}, can be combined to potentially solve various tasks, compositionally. None of our modules are fine-tuned explicitly for the target problems, placing us in a zero-shot setting. 
Moreover, \approach\ does not only rely on an exploration policy based entirely on \ac{llm} output, as in \cite{zhou2024navgpt,zhou2025navgpt,dorbala2023can}. Instead, it combines a frontier exploration policy with \ac{llm} priors and memory guidance to enhance navigation performance.
To achieve this, we use a pre-trained PointGoal policy as the base waypoint navigation module of \approach\ and extend the exploration policy introduced in \cite{yokoyama2024vlfm} by equipping the agent with a memory mechanism stored as a feature vector for each pixel of the map. 

\paragraph{Modular vision and program composition for visual tasks.}
Neural Module Networks (NMN)~\cite{nmn2016, hu2017learning} introduced modular and compositional methodologies for visual question answering (VQA). NMNs integrate neural modules into an end-to-end differentiable network; the approach originally relied on pre-existing parsers~\cite{nmn2016}, whereas more recent methods~\cite{hu2017learning, Johnson2017InferringAE, Hu2018StackNMN} have evolved to learn the layout generation model concurrently with the neural modules, employing reinforcement learning~\cite{williams1992reinforce} and weak supervision. Stack-NMN~\cite{Hu2018StackNMN} extends N2NMN by transitioning from discrete to soft layout generation, incorporating a weighted average of predictions from all modules at each step, determined by a layout generation network.

Recently, \cite{visprog} introduced VisProg, a framework that offers two key advantages over NMNs. Firstly, it constructs high-level programs that invoke state-of-the-art neural models and Python functions at intermediate steps, diverging from the conventional approach of generating end-to-end neural networks. This design facilitates the integration of symbolic, non-differentiable modules. Secondly, it capitalizes on the in-context learning ability of large language models (LLMs)~\cite{gpt} to generate programs. This is achieved by prompting the LLM with a natural language instruction (or a visual question or a statement to be verified), along with a few examples of similar instructions and their corresponding programs. This approach eliminates the need to train specialized program generators for each task. 
Almost concurrently, ViperGPT~\cite{vipergpt} presented a similar approach to VisProg, but in this case the model directly generates Python code, instead of composing pre-defined modules and routines.
In contrast, in our framework, the \ac{llm} is provided only with high-level knowledge of the functions of each primitive, which is encoded directly in the primitive's name. Therefore, it is built similarly to VisProg, as directly creating code to solve EAI tasks is a much harder problem than composing pre-defined primitives.
It is noteworthy that both VisProg and ViperGPT were tested on images, while our model is specifically designed for embodied agents operating in an interactive environment, relying on sensor-based perception.


\section{Method}
\label{sec:method}
\long\def\/*#1*/{}

\begin{figure*}[!t]
\begin{center}
\includegraphics[width=.95\textwidth]{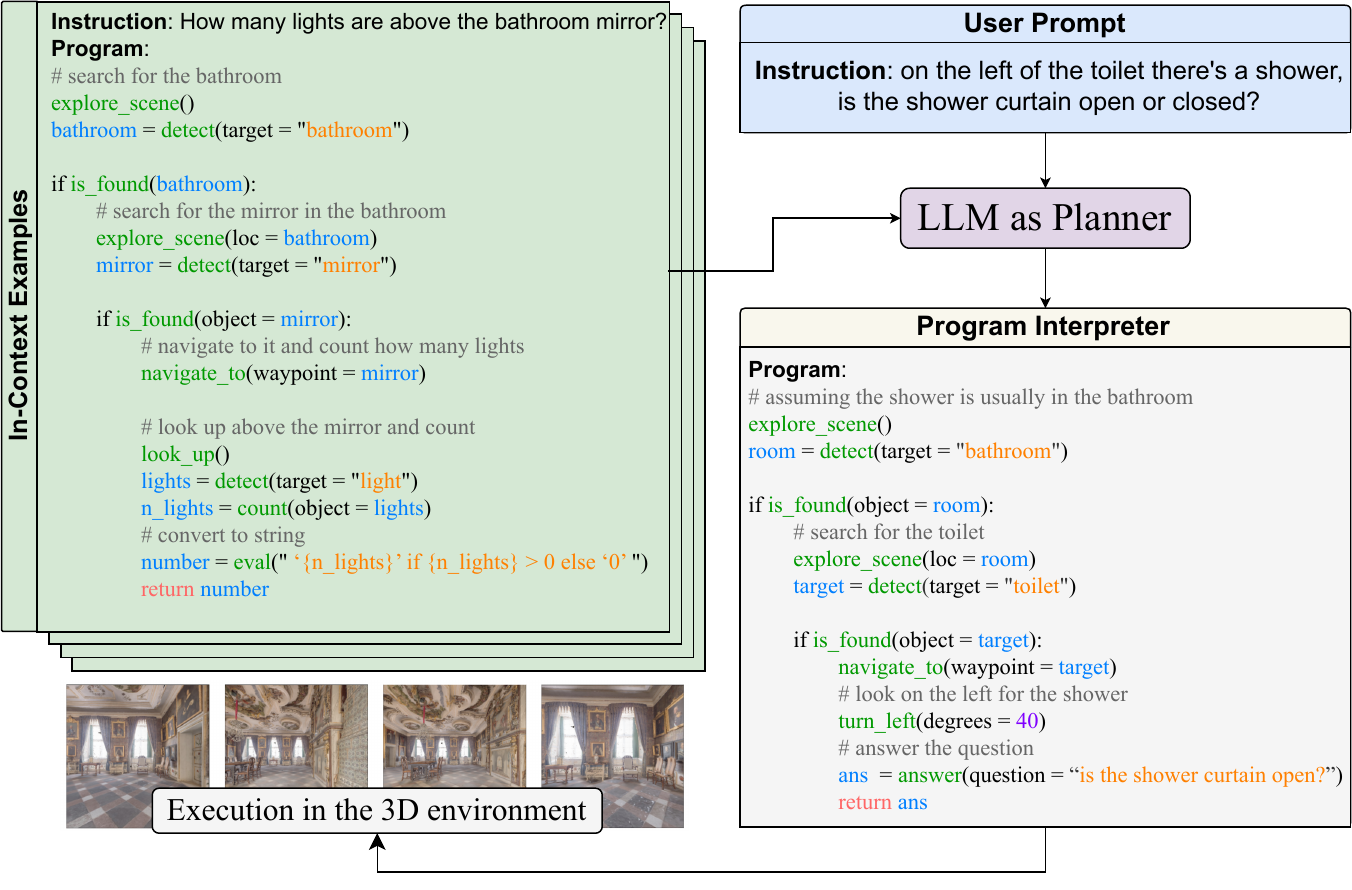}
\caption{\textbf{Overview of the program generation in \approach.} Given few ``in-context examples", the \ac{llm} provide a detailed sequence of steps to be executed by the agent in the given environment. The \ac{llm} is instructed to comment its output to allow for explainability.}
\vspace{-10pt}
\label{img:method_img}
\end{center}
\end{figure*}

Neuro-symbolic approaches offer the possibility to address a broad range of diverse complex tasks efficiently. Systems like VisProg~\cite{visprog} operate on images and have demonstrated excellent results in zero-shot settings. Understanding how a similar approach works in an action-oriented scenario is key to enable robots to navigate autonomously in novel environments.
Therefore, we develop this paradigm further into EAI, where the integration with action execution adds several challenges. 
Inspired by \cite{visprog}, given an input prompt, we rely on a \ac{llm} to generate synthetic pseudo-programs that can be executed by the embodied agent in the environment. 
Our \approach\ framework is therefore able to generalize to new tasks, without any direct training on task-specific datasets, thus effectively mitigating the heavy computational training costs inherent to embodied navigation tasks. The procedure that enables generating these new programs is based on ``in-context examples'' that are fed to the \ac{llm}, alongside a natural language instruction. An outline of \approach\ is shown in Figure~\ref{img:method_img}.

\vspace{-10pt}
\paragraph{\approach\ Interpreter.}
The first key component of the proposed framework is referred to as ``Program Interpreter''.
It comprises visual recognition modules that can be used by the agent to extract the semantics of the scene, as well as to provide an understanding of the visual context. 

\begin{figure}[!t]
\begin{center}
\includegraphics[width=0.48\textwidth]{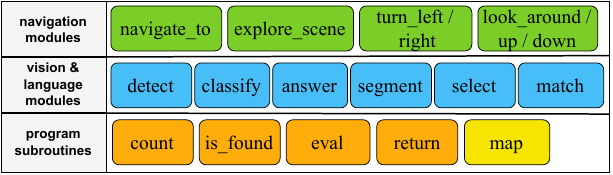}
\end{center}
\vspace{-15pt}
\caption{\textbf{Overview of \approach\ modules.} Modules span a variety of inputs and outputs. Orange modules use Python subroutines, while blue modules use pre-trained computer vision models (similarly to \cite{visprog}). The \textit{navigate\_to} and \textit{explore\_scene} modules, in green, both implement our foundational PointNav module; however, only \textit{explore\_scene} integrates the memory mechanism.}
\label{img:navprog_modules}
\end{figure}

As shown in Figure~\ref{img:teaser}, users can ask questions or give specific tasks to the agent; the natural language prompt is then processed by the \ac{llm} (in our implementation GPT-4o~\cite{gpt}), which serves as a planner and outputs a step-by-step executable program. 
To ensure the \ac{llm} delivers a reasonable output, it is fed with $15$ ``in-context examples'' across diverse tasks. This enables the \ac{llm} to make use of its reasoning capabilities effectively, identifying the most suitable planning for the required task. 
Moreover, the examples remain the same regardless of the task identified in the prompt; it is the \ac{llm}'s responsibility to output the specific program target for the given question.
Programs use a higher level of abstraction than previous modular attempts such as Neural Module Networks (NMN)~\cite{nmn2016, hu2017learning}. Each program is constructed as a sequence of primitives (e.g., detect, answer, match, etc.) that invoke corresponding \approach\ modules. These modules are either powered by pre-trained state-of-the-art vision models or implemented as simple Python subroutines (e.g., count, is\_found, eval, etc.), with additional navigation-specific modules designed to ``steer'' the agent's movements (e.g. navigate\_to, return, turn, etc.). Figure~\ref{img:navprog_modules} provides a comprehensive list of all the currently implemented modules.

Figure~\ref{img:method_img} shows an \ac{eqa} example for the user's question: \emph{``on the left of the toilet there's a shower, is the shower curtain open or closed?''}. The \ac{llm}, which acts as a planner, transforms the user's initial prompt into navigation subtasks by leveraging its reasoning abilities and powerful priors. This decomposition enables the model to break down complex queries into more manageable steps, guiding the agent's navigation process effectively.
Each line of the generated program corresponds to a module serving a specific task. The agent must then execute the generated program. 
The \ac{llm} is also guided to comment on its steps directly within the generated pseudocode.

All the given primitives are equipped with methods to: \emph{i})~\textbf{parse} lines in order to extract input argument names and values, as well as the output variable name; \emph{ii})~\textbf{execute} the module in the environment, and update the program state with the output variable name and value. 
The outputs for each step and the comments in the generated pseudocode can also be used to better understand what is happening under the hood and why the \ac{llm} generated a specific line, thus enabling a good interpretability of the system behavior. 

\approach\ modules utilize open-source software and models, readily downloadable from the web. They can also be effortlessly updated with newer, more efficient models as they become available.

\vspace{-10pt}
\paragraph{Navigation Module.} \label{navigation_module}
In order to navigate the environment, we define a module employing a PointGoal navigation agent as our foundational module~\cite{raychaudhuri2024mopa, yokoyama2024vlfm}. This model achieved nearly perfect PointGoal performance, both in terms of success ($\sim99\%$) and efficiency (a forward pass that takes a fraction of a second) on standard datasets~\cite{ramakrishnan2021habitat, chang2017matterport}. 
The agent starts the exploration of the environment until it locates its target goal. Once the target is identified, the focus of exploration transitions to reaching the designated goal. This PointNav policy exclusively uses the egocentric depth image and the robot’s relative distance and heading towards the desired goal point as input.

\vspace{-10pt}
\paragraph{Exploration Policy.} \label{exploration_policy}

Exploration is performed using the policy outlined in \cite{yokoyama2024vlfm}. This method builds occupancy maps based on depth observations to identify exploration frontiers. It further leverages RGB observations and a pre-trained vision-language model (BLIP2~\cite{li2023blip}) to generate a language-grounded ``value map''. This value map is then used to guide the exploration, facilitating an efficient search for instances of a given target. Notably, as in \cite{yokoyama2024vlfm}, the agent performs a $360^{\circ}$ turn at the start of navigation to initialize frontiers.

We extend this policy for sequential goals by incorporating a memory mechanism, represented as a ``feature map'' in which each pixel of the value map is encoded as a vector and updated at each step. 
This feature map then updates the language-grounded original value map when a new target is specified, by calculating the cosine similarity between each pixel's vector and the new target embeddings (either from text or image). 
We sample the highest value in the updated value map; if this value exceeds a predefined threshold, indicating the agent has previously encountered and ``remembers'' the target's location, the agent navigates to it, supporting lifelong navigation.
The process is also computationally efficient, as the feature map similarity calculations are performed only if the target changes, rather than at each step during navigation. 


\subsection{Navigation Tasks with \approach} \label{navigation_tasks_with_navprog}
\approach\ leverages pre-trained multimodal vision-language and vision-only models as foundational components to extract semantic information from the scene, making it well-suited for several Embodied AI tasks. 

(Open-set) ObjectGoal Navigation~\cite{batra2020objectnav, yokoyama2024hm3d}, in particular, emerges as a suitable testbed to evaluate the efficacy of our method. A key module to tackle this problem involves utilizing an object detector (Owlv2 in our implementation~\cite{minderer2024scaling} or DETR~\cite{carion2020end} for target objects that fall within the COCO classes~\cite{lin2014microsoft}) to identify objects within the image. 
Subsequently, navigation towards the detected objects is facilitated by waypoints, leveraging depth distance calculation to guide the agent effectively. For simplicity, we use the center of the bounding box to determine the target waypoint.
Hence, our approach effectively addresses this task without requiring prior knowledge of target labels for previously encountered objects, making it suitable for real-world applications.

\approach\ also integrates a specialized module for matching target objectives with ongoing visual observations in image-based scenarios, resulting in accurate navigation towards an instance image. To evaluate our agent's performance in this context, we use SuperGlue network~\cite{superglue}, optimized for indoor environments (i.e. using hyperparameters recommended in the respective paper).
We rely on SuperGlue because of its real-time matching capabilities and effortless integration into Embodied systems.

\approach\, being task-agnostic, can process any type of navigation target, regardless of the order in which it is presented. In this context, the GOAT benchmark~\cite{khanna2024goat} is well-suited for evaluating our system, as it operates in scenarios where targets are specified through images, text, or descriptive phrases, provided in random sequence. It also heavily relies on memory to find previously seen targets when sequential goals are given.

An agent should also be able to answer user queries like \emph{``can you check if the kitchen table is clean?''}. Hence, \ac{eqa} serves as an excellent benchmark to assess \approach's capabilities and robustness. \ac{eqa} consists of three phases: understanding the semantic structure of the prompt, locating the target object(s), and analyzing the visual semantics to generate an accurate response. To this end, \approach\ integrates a specific \textit{answer} module, relying on BLIP2~\cite{li2023blip} as its core foundation.


\section{Experiments}
\label{sec:exp}

\approach\ provides a flexible framework that can be applied to various embodied navigation problems. 
We evaluate our approach on three popular tasks that require a wide range of capabilities, including efficient environment exploration, path planning, scene and context understanding, and image similarity comparison.

\subsection{Experimental Setup}

\paragraph{Agent Configuration.}
Prior research on visual navigation commonly uses different agent configurations depending on the considered task~\cite{khanna2024goat, yokoyama2024hm3d, das2018embodied}. The configuration typically employed for \textsc{(Open-Set) ObjNav} closely resembles that of a LoCoBot, with an agent height of $0.88m$, a radius of $0.18m$, and a single $640\times480$ RGB sensor with a $79^{\circ}$ hfov, positioned $0.88m$ above the ground. 


In the context of GOAT Benchmark, some of the default settings differ, featuring a camera situated $1.31m$ above the ground. The agent has a height of $1.41m$ and is given $360\times640$ RGB images. 
Both the above tasks' configurations use a step size of $0.25m$ and a left and right turning angle of $30^{\circ}$.
Lastly, the settings for \ac{eqa} mirror those of PointGoal Navigation as specified in the Habitat-Lab configuration files. All other settings remain consistent with those of \cite{das2018embodied}.

\paragraph{Datasets and Evaluation Protocol.}
We assess the \approach\ performance across task-specific datasets using the Habitat simulator~\cite{savva2019habitat}. As the system is not trained in any way (except for the PointGoal Navigation model, which has been pre-trained on the training set of HM3D, never used in our tests), each scene within episodes is novel to the agent. Therefore, the entire validation set can be categorized as \textit{``unseen''}.
We now describe in detail the tasks and datasets used:

\begin{itemize}

    \item \textsc{Open Vocabulary Object Navigation (OVON)~\cite{yokoyama2024hm3d}}: a large-scale benchmark featuring over 15,000 annotated household objects across $379$ categories, derived from real-world 3D scans~\cite{ramakrishnan2021habitat}. 
    The agent is initialized at a random location within a scene and tasked with navigating to a target object category within a time limit of $500$ steps. 
    To ensure comparability with other \ac{sota} methods, we evaluate our approach across all episodes included in the \textit{``val unseen''} set.

    \item \textsc{Multi-Modal Lifelong Navigation (GOAT-Bench)~\cite{khanna2024goat}}: an agent is tasked with sequentially navigating to five to ten target objects identified by category names, descriptions, or images. Each target represents a subtask executed within an open-vocabulary framework that spans over $312$ categories. The agent is required to reach a goal within a specified time constraint and is assigned new targets upon the completion of each subtask. We evaluate our approach across all episodes included in the \textit{``val unseen''} set.

    \item \textsc{Embodied Question Answering (OpenEQA)~\cite{majumdar2024openeqa}}: containing over 1,600 question-answer pairs sourced from more than 180 real-world environments and scans~\cite{ramakrishnan2021habitat, Dai_2017_CVPR}. It is divided into two task categories: Episodic Memory-EQA (EM-EQA) and Active-EQA (A-EQA). Our focus is on the latter, in which the agent must autonomously navigate and answer questions within a time constraint of $500$ steps. Questions span over seven categories, including world knowledge, attribute recognition, spatial reasoning, and object localization. Examples of questions from the episodes include: \textit{``Is the microwave door propped open?''} and \textit{``What is left of the kitchen pass-through?''}. We evaluate our approach across all episodes belonging to the A-EQA task category. 

\end{itemize}

\paragraph{Evaluation Metrics.}
\label{sec:metrics}
In all tasks, we use the standard metrics as in prior works~\cite{batra2020objectnav, majumdar2024openeqa, das2018embodied, yokoyama2024hm3d}. Namely:
\begin{itemize}
    \item \emph{Success Rate} (SR): measures the ratio of episodes where the agent succesfully reached its target (Open ObjNav, GOAT-Bench).
    \item \emph{Success weighted by Path Length} (SPL): measures the optimality of the path taken by the agent \wrt the optimal path (Open ObjNav, GOAT-Bench).
    \item \emph{Distance to Goal} (DTG): measures the average distance to goal of the agent at the end of the episode (Open ObjNav, GOAT-Bench).
    \item \emph{Score} (LLM-Match): an \ac{llm} compares the ground-truth $GT_{i}$ answers with model output $A_{i}$ given a question $Q_{i}$ and assigns a score $\sigma_{i}$ on a scale of 1 to 5. On this scale, 1 indicates an incorrect response, 5 represents a correct response, and intermediate values reflect varying levels of similarity. For example, the answer to the question \textit{``What color is the bed''} could be correctly answered either as \textit{``white''} or \textit{``the bed is white''}. The final results aggregation is as follows:
    \begin{equation}
        S = \frac{1}{N} \sum_{i}^{N} \frac{\sigma_{i} -1}{4} \times 100\%
        \label{eq:llm_score}
    \end{equation}
    \item \emph{Answer Accuracy}: the average accuracy of the answers provided by the agent for the EQA task. This metric, used in \cite{das2018embodied}, is presented with the related results in the supplementary material.
\end{itemize}

\paragraph{Implemented Modules.}

\approach\ contains descriptive and easily understandable names for the modules, arguments, and variables to facilitate the \ac{llm} comprehension of the input and output. Figure \ref{img:navprog_modules} showcases a list of all available modules. Each module exclusively outputs pre-defined object variables to the next one, enabling the possibility to monitor the progression of the agent at each step. Therefore, dramatically enhancing the explainability of failure cases, as it allows for a deep examination of the agent's decision-making process through its interactions with the environment. Figure \ref{img:failure_analysis} provides a general overview of the failure analysis in \ac{eqa} task. 
As presented in Section \ref{navigation_module}, ``navigate\_to'' and ``explore\_scene'' modules are specifically engineered to navigate the environment, leveraging inputs from depth and RGB sensors as well as the PointGoal GPS + compass sensors. If this sensor is not available for a specific task, it can be derived from the current agent pose and the exploration goal location. The agent is provided with the actions: \textit{Forward}, \textit{Move Left}, \textit{Move Right}, \textit{Look Up}, \textit{Look Down} and \textit{Stop}. 

For the ``detect'' module, we employ Owlv2~\cite{minderer2024scaling} object detector for general classes and use DETR~\cite{carion2020end} for categories within the COCO classes~\cite{lin2014microsoft}. Notably, in synthetic 3D scenes, we noticed numerous false positives. To address the problem, bounding boxes produced by the detector are forwarded to a ``classify'' module. Within this module, each detection undergoes classification to differentiate between categories of similar instances. For example, the class \textit{``chair''} encompasses diverse subcategories such as \textit{``armchair''}, \textit{``couch''}, and \textit{``other''}. Subclasses are outputted under the hood directly by the \ac{llm} requiring no human annotation. We utilize a CLIP-based classifier in its default configuration~\cite{radford2021learning}. 

Embodied Question Answering (\ac{eqa}) task uses an ``answer'' module based on BLIP2~\cite{li2023blip}, capable of performing various multi-modal tasks, including Visual Question Answering, Image-Text retrieval, and Image Captioning. Notably, the module is also used when goals are specified through images, to capture the class of the input image before starting the navigation. Furthermore, for image-based tasks we exploit a ``match'' module to evaluate the similarity between the agent's observation and the instance image target. This module performs feature matching between the two, as described in Section \ref{navigation_tasks_with_navprog}, to aid in navigation. Hence, it assists the agent in determining whether what it perceives as the current target is indeed the correct goal.

\subsection{Experimental Results}


\begin{figure}[t]
\centering
\includegraphics[width=0.48\textwidth]{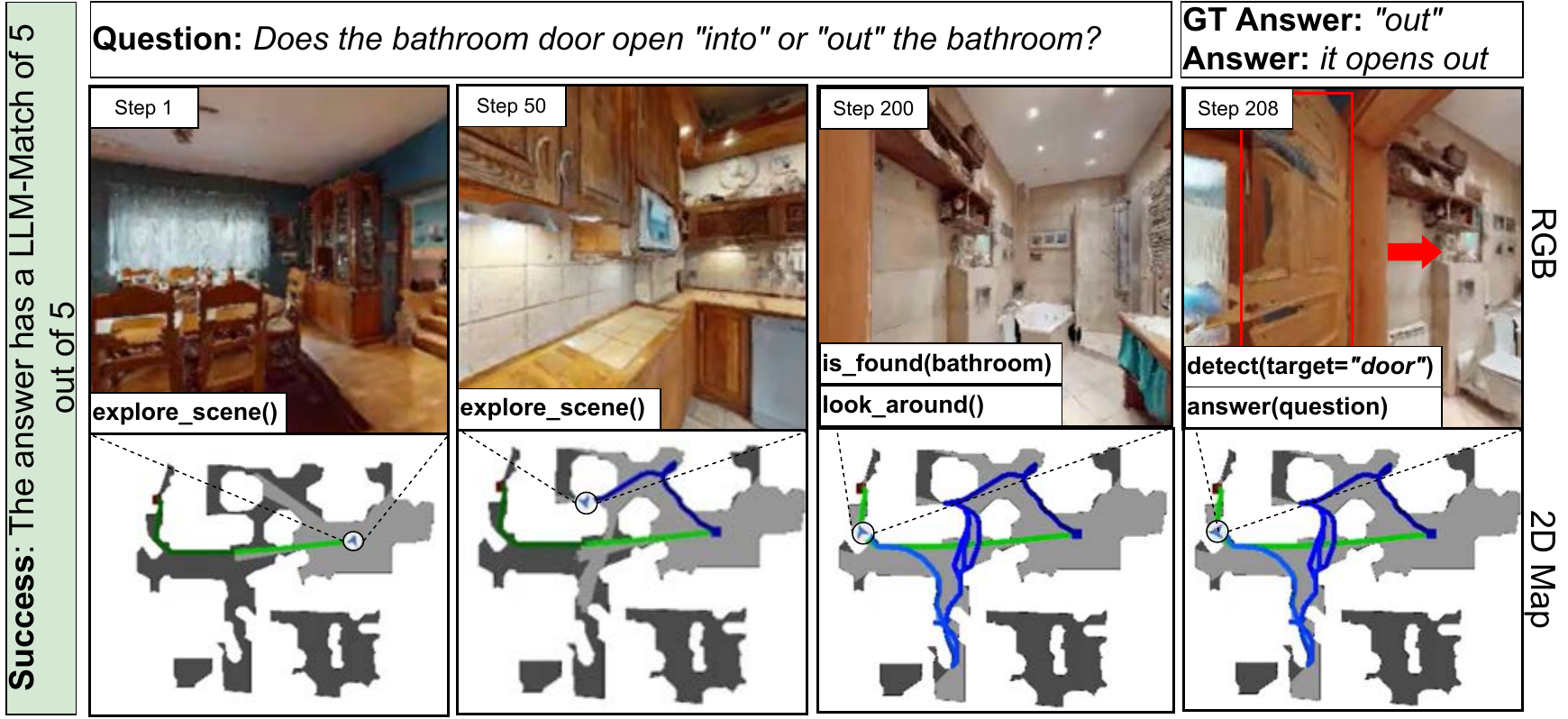}
\includegraphics[width=0.48\textwidth]{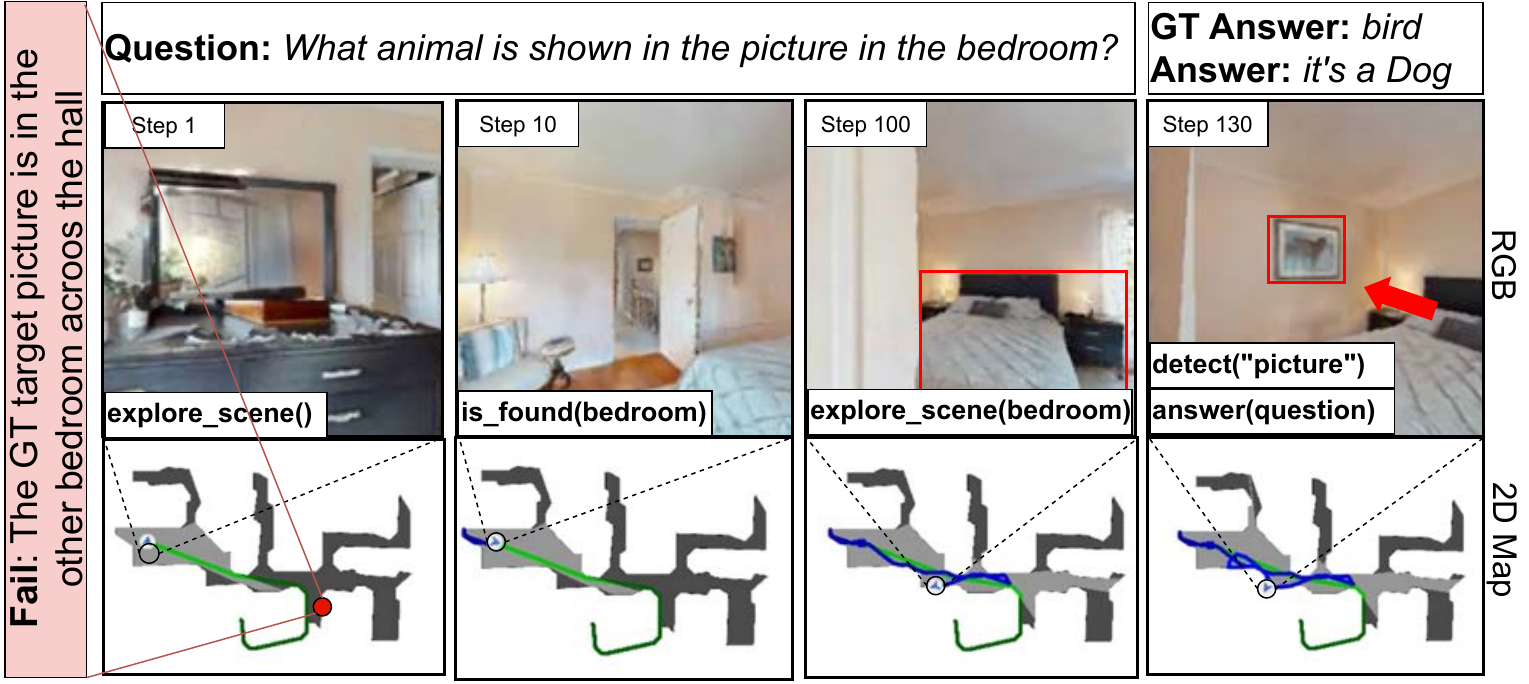}
\caption{\textbf{Examples from OpenEQA}~\cite{majumdar2024openeqa}. The top section illustrates a successful episode where \approach\ is able to understands the input query, correctly specifying the sequential targets. The lower section illustrates a failure caused by overly general directions from the \ac{llm}, which \approach\ struggled to resolve.}
\label{fig:open_eqa_examples}
\end{figure}


\paragraph{Open-set ObjectGoal Navigation}
We evaluated our method in the standard Open-Set OBJNAV setting, with targets sampled directly from objects in the scene. As shown in Table \ref{tab:ovon_results}, our approach achieves performance on par with state-of-the-art methods (rows 4-5) on the validation-unseen split both in SR ($35.5\%$) and SPL ($19.5\%$). Instead of focusing solely on numerical results, these findings highlight \approach\ ’s flexibility in solving navigation tasks in unfamiliar environments, underscoring the potential of our approach within the OBJNAV framework.

\setlength{\tabcolsep}{4pt}
\begin{table}[t]
\centering
\begin{tabular}{lll}
\toprule
 & \multicolumn{2}{c}{\textbf{VAL UNSEEN}} \\ 
\cmidrule(lr){2-3}
\textbf{Method}   & SR ($\uparrow$) & SPL ($\uparrow$) \\
\midrule
\noalign{\smallskip}
RL  \cite{yokoyama2024hm3d}        & 18.6 \begin{footnotesize}$\pm 0.3 $\end{footnotesize}        & 7.5 \begin{footnotesize}$\pm 0.2$\end{footnotesize} \\
BCRL \cite{yokoyama2024hm3d}    & 8.0 \begin{footnotesize}$\pm 0.2 $\end{footnotesize}        & 2.8 \begin{footnotesize}$\pm 0.1$\end{footnotesize} \\
DAgRL \cite{yokoyama2024hm3d}     & 18.3 \begin{footnotesize}$\pm 0.3 $\end{footnotesize}         & 7.9 \begin{footnotesize}$\pm 0.1$\end{footnotesize}  \\
\noalign{\smallskip}
\hline
\noalign{\smallskip}
$^{*}$VLFM \cite{yokoyama2024hm3d, yokoyama2024vlfm}        & 35.2         & 19.6 \\
DAgRL+OD \cite{yokoyama2024hm3d}  & \textbf{37.1 \begin{footnotesize}$\pm 0.2$ \end{footnotesize}}        & \textbf{19.9 \begin{footnotesize}$\pm 0.3$\end{footnotesize}} \\
\noalign{\smallskip}
\hline
\noalign{\smallskip}
\textbf{\approach\ (ours)}  & 35.5 \begin{footnotesize}$\pm 0.3$ \end{footnotesize}       & 19.5 \begin{footnotesize}$\pm 0.3$ \end{footnotesize} \\
\bottomrule
\end{tabular}
\caption{\textbf{HM3D-OVON}. Performance comparison of different methods on VAL UNSEEN split. $^{*}$ deterministic method.}
\label{tab:ovon_results}
\end{table}
\setlength{\tabcolsep}{1.4pt}

\paragraph{Embodied Question Answering.} \label{sec:eqa} 
In \textsc{\ac{eqa}}, the agent is queried with a natural language question and must autonomously explore the environment and gather information in order to answer accordingly. In recent literature, this task shifted from being regarded as a purely classification problem aimed at determining the most suitable answer from a set of pre-defined possibilities (i.e., class labels) to an open-vocabulary benchmark where the agent must answer with natural language~\cite{das2018embodied, majumdar2024openeqa}.
We compare \approach\ results, using the score from Eq. \ref{eq:llm_score}, against other zero-shot approaches. Table \ref{tab:open_eqa_results} shows that our method ranks second and closely matches the performance of the leading SoA method, with a gap of less than $1\%$. Notably, these approaches~\cite{majumdar2024openeqa} utilize \acp{llm} to interpret scene objects step-by-step (rows 3-4), but they do not leverage \acp{llm} to effectively guide navigation towards specific targets. Hence, our approach demonstrates that SoA results can still be achieved by pre-planning a path that prioritizes relevant areas for a given question. 
Moreover, the results indicate that human agents achieve a significantly higher score of $85\%$, which remains well above the performance achieved by using large language models (LLMs), particularly in open-world settings. This gap underscores the challenges LLMs face in handling the complexity and variability of open-world scenarios compared to human agents.

\setlength{\tabcolsep}{4pt}
\begin{table}[t!]
\centering
\begin{tabular}{ll}
\toprule
\noalign{\smallskip}
\textbf{Method} & Score ($\uparrow$) \\ 
\noalign{\smallskip}
\hline
\noalign{\smallskip}
\rowcolor[gray]{0.9} Human Agent \cite{majumdar2024openeqa} & 85.1 \begin{footnotesize} $\pm$ 1.1 \end{footnotesize} \\ 
\noalign{\smallskip}
\hline
\noalign{\smallskip}
Blind LLMs \cite{majumdar2024openeqa} & 35.5 \begin{footnotesize} $\pm$ 1.7 \end{footnotesize}\\ 
Socratic LLMs w/ Frame Captions \cite{majumdar2024openeqa}  & \textbf{38.1} \begin{footnotesize} $\pm$ 1.8 \end{footnotesize}\\
Socratic LLMs w/ Scene-Graph Captions \cite{majumdar2024openeqa}  & 34.4 \begin{footnotesize} $\pm$ 1.8 \end{footnotesize}\\ 
\noalign{\smallskip}
\hline
\noalign{\smallskip}
\textbf{\approach\ (ours)}  & 37.2 \begin{footnotesize} $\pm$ 1.8 \end{footnotesize}\\ 
\bottomrule
\end{tabular}
\caption{\textbf{OpenEQA results.} Comparison of \approach\ with the SoA on the OpenEQA dataset. All methods involve zero-shot approaches.}
\label{tab:open_eqa_results}
\vspace{-15pt}
\end{table}
\setlength{\tabcolsep}{1.4pt}

\begin{figure}[t]
\centering
\includegraphics[width=0.48\textwidth]{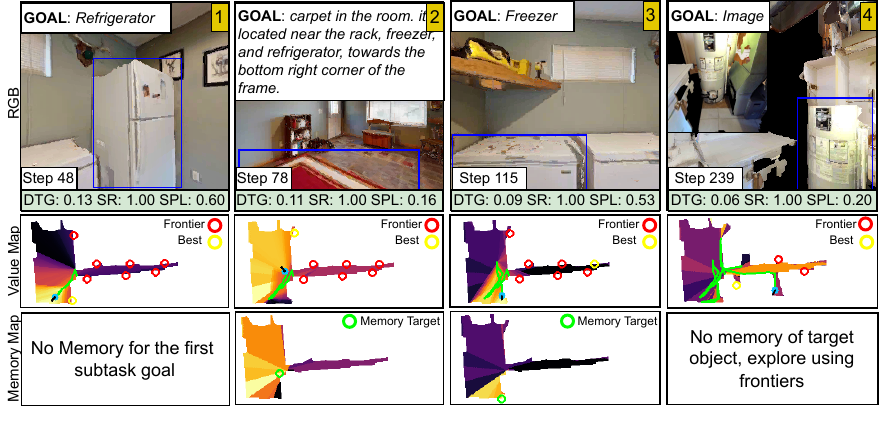}
\caption{\textbf{Multi-Modal Lifelong Navigation Success Example.} (top) RGB observation of the target during STOP action (step $t_{i+ steps}$). (middle) Value map for the specific target recomputed from the memory map (step $t_{i + steps})$. (bottom) Memory map after target changes (step $t_{i}$). }
\label{fig:goat_success_example}
\vspace{-15pt}
\end{figure}

\paragraph{GOAT.}\label{sec:goat} 
We assess our method within the GOAT-Bench setting, where the agent is spawned randomly and tasked with sequential targets. Each goal can be specified either by its category name, a description, or an image (e.g. the input could be \textit{``gas boiler''}, \textit{``the gas boiler on the corner of the room. The gas boiler is located on the left of the washing machine and freezer''} or an image of the \textit{``gas boiler''}).
Table \ref{tab:goat_results} compares our approach to the methods in \cite{chang2023goat}. \approach\ significantly outperforms other state-of-the-art techniques (rows 1-4), achieving a $+2.6\%$ gain in the success metric. Additionally, it ranks second in navigation efficiency, with only a slight $-0.7\%$ difference from the top-performing method. We primarily attribute these results to the \ac{llm}'s ability to accurately transform target descriptions into a sequence of selected goals, rather than relying solely on potentially ``noisy'' text embeddings. Moreover, the use of a memory mechanism is essential for the task (Fig. \ref{fig:goat_success_example}); however, if the memorized target object is incorrect, the path efficiency can significantly decrease.

\setlength{\tabcolsep}{4pt}
\begin{table}[t!]
\centering
\begin{tabular}{lll}
\toprule
 & \multicolumn{2}{c}{\textbf{VAL UNSEEN}} \\ 
\cmidrule(lr){2-3}
\textbf{Method}   & SR ($\uparrow$) & SPL ($\uparrow$) \\
\midrule
\noalign{\smallskip}
SenseAct-NN Skill Chain \cite{khanna2024goat} & 29.5 & 11.3 \\
SenseAct-NN Monolitic  \cite{khanna2024goat} & 12.3          & 6.8  \\
\noalign{\smallskip}
\hline
\noalign{\smallskip}
Modular GOAT            \cite{khanna2024goat} & 24.9          & \textbf{17.2} \\
Modular Clip on Wheels  \cite{khanna2024goat} & 12.3          & 10.4 \\
\noalign{\smallskip}
\hline
\noalign{\smallskip}
\textbf{\approach\ (ours)}      & \textbf{32.1}        & 16.5 \\
\bottomrule
\end{tabular}
\caption{\textbf{GOAT-Bench}. Performance comparison of different methods on the VAL UNSEEN split.}
\label{tab:goat_results}
\end{table}
\setlength{\tabcolsep}{1.4pt}

\subsection{\approach\ Failure analysis} \label{sec:failure_analysis}
\begin{figure}[t]
\begin{center}
\includegraphics[width=0.45\textwidth]{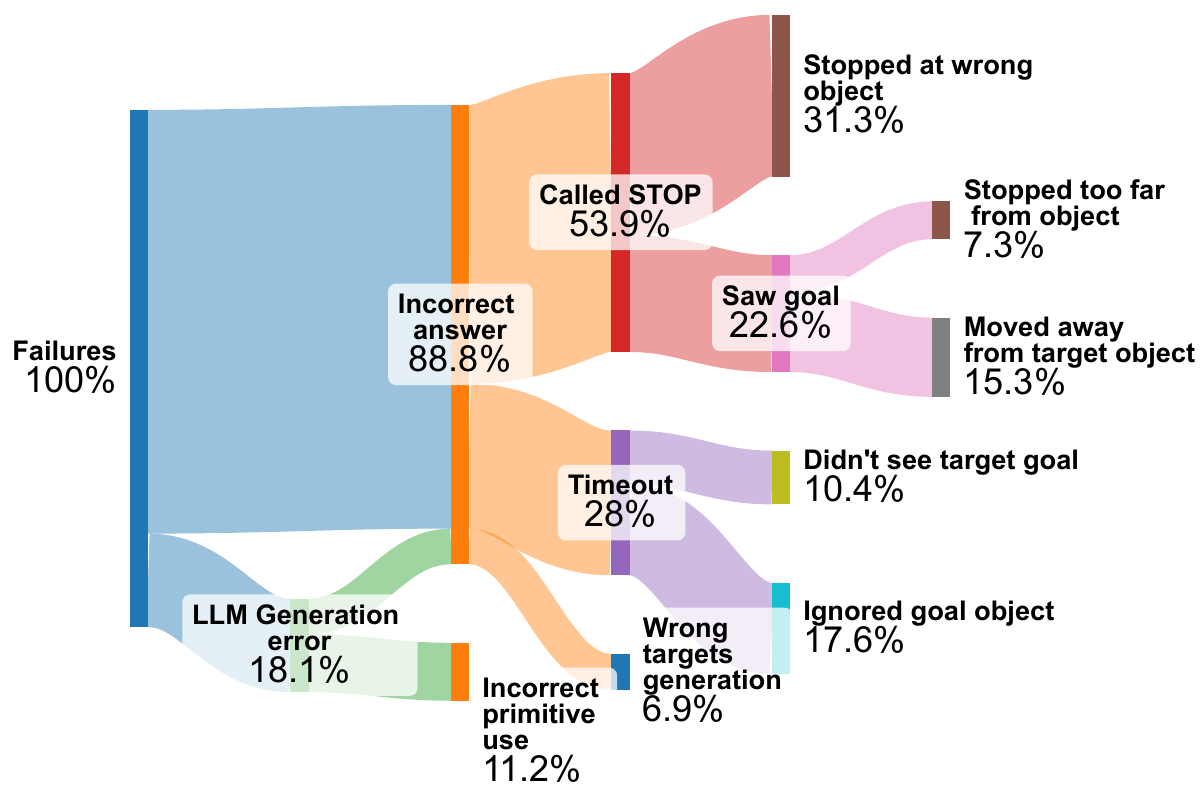}
\caption{\textbf{Causes of system failure in \ac{eqa} tasks.} The analysis was manually conducted on a significant sample of failed episodes, facilitated by the explanaibility of \approach\ . Answers were deemed as ``incorrect'' if they received a score below three.
}
\label{img:failure_analysis}
\end{center}
\vspace{-10pt}
\end{figure}

Due to the high explainability of our system, we conducted an analysis of the failures encountered during \ac{eqa} episodes (Fig. \ref{img:failure_analysis}). We extracted a significant subsample for the task and manually classified the instances where the model failed.
We considered scores below 3 as failures, i.e. incorrect answers. We observed that the main cause of failure can be attributed to the ``detect'' module (\textit{stopped at wrong object} or \textit{Ignored goal object} in the image). Furthermore, the exploration policy appears to perform well given the targets produced by the \ac{llm}, as it fails in only $\sim10\%$ of the cases (labeled as: \textit{didn't see target goal} in the image). The \ac{llm} generates incorrect pseudo-code around $18\%$ of the time, with $6.9\%$ failures leading to ambiguous or incorrect targets, and $11.2\%$ due to incorrect primitive ordering or usage which instead leads directly to episode failure. 
Notably, only $11.2\%$ of errors are actually due to the \ac{llm} generating incorrect code, while the remaining is attributable to prompt-related issues.

\section{Conclusion}
\label{sec:conclusion}
We conducted a systematic analysis of our compositional approach,
offering valuable insights into the future of modular, neuro-symbolic systems. 
Moreover, \approach\ highlights the versatility of simple PointGoal Navigation agents equipped with specific task-oriented modules, yielding promising results in zero-shot scenarios across all considered tasks. 
\approach\ effectively handles diverse multi-modal prompts, following instructions to complete tasks, underscoring the potential of \acp{llm} in robotic navigation. Its integrated memory supports lifelong navigation, suggesting improved capabilities with increased exploration.
However, it may struggle with overly complex or ambiguous prompts, which can limit the \ac{llm}'s ability to identify the correct primitives or targets. 
Future work could extend this system to tasks such as Visual Language Navigation (VLN) and explore the use of open-source \acp{llm}. 
Additionally, refining the memory mechanism—particularly through improved sampling strategies for identifying high-value pixels linked to objects' memory—could further enhance the system’s effectiveness.

{
 \small
 \bibliographystyle{ieeenat_fullname}
 \bibliography{main}

\begin{thebibliography}{56}
\providecommand{\natexlab}[1]{#1}
\providecommand{\url}[1]{\texttt{#1}}
\expandafter\ifx\csname urlstyle\endcsname\relax
  \providecommand{\doi}[1]{doi: #1}\else
  \providecommand{\doi}{doi: \begingroup \urlstyle{rm}\Url}\fi

\bibitem[Anderson et~al.(2018)Anderson, Wu, Teney, Bruce, Johnson, S{\"u}nderhauf, Reid, Gould, and van~den Hengel]{anderson2018vision}
Peter Anderson, Qi Wu, Damien Teney, Jake Bruce, Mark Johnson, Niko S{\"u}nderhauf, Ian Reid, Stephen Gould, and Anton van~den Hengel.
\newblock Vision-and-language navigation: Interpreting visually-grounded navigation instructions in real environments.
\newblock In \emph{Proc. of the IEEE/CVF Conference on Computer Vision and Pattern Recognition (CVPR)}, 2018.

\bibitem[Andreas et~al.(2016)Andreas, Rohrbach, Darrell, and Klein]{nmn2016}
Jacob Andreas, Marcus Rohrbach, Trevor Darrell, and Dan Klein.
\newblock Neural module networks.
\newblock In \emph{Proc. of the IEEE/CVF Conference on Computer Vision and Pattern Recognition (CVPR)}, 2016.

\bibitem[Batra et~al.(2020)Batra, Gokaslan, Kembhavi, Maksymets, Mottaghi, Savva, Toshev, and Wijmans]{batra2020objectnav}
Dhruv Batra, Aaron Gokaslan, Aniruddha Kembhavi, Oleksandr Maksymets, Roozbeh Mottaghi, Manolis Savva, Alexander Toshev, and Erik Wijmans.
\newblock {ObjectNav} revisited: On evaluation of embodied agents navigating to objects.
\newblock \emph{arXiv preprint arXiv:2006.13171}, 2020.

\bibitem[Bing et~al.(2023)Bing, Koch, Yao, Huang, and Knoll]{bing2023meta}
Zhenshan Bing, Alexander Koch, Xiangtong Yao, Kai Huang, and Alois Knoll.
\newblock Meta-reinforcement learning via language instructions.
\newblock In \emph{2023 IEEE International Conference on Robotics and Automation (ICRA)}, pages 5985--5991. IEEE, 2023.

\bibitem[Brown et~al.(2020)Brown, Mann, Ryder, Subbiah, Kaplan, Dhariwal, Neelakantan, Shyam, Sastry, Askell, Agarwal, Herbert-Voss, Krueger, Henighan, Child, Ramesh, Ziegler, Wu, Winter, Hesse, Chen, Sigler, Litwin, Gray, Chess, Clark, Berner, McCandlish, Radford, Sutskever, and Amodei]{gpt}
Tom~B. Brown, Benjamin Mann, Nick Ryder, Melanie Subbiah, Jared Kaplan, Prafulla Dhariwal, Arvind Neelakantan, Pranav Shyam, Girish Sastry, Amanda Askell, Sandhini Agarwal, Ariel Herbert-Voss, Gretchen Krueger, Tom Henighan, Rewon Child, Aditya Ramesh, Daniel~M. Ziegler, Jeffrey Wu, Clemens Winter, Christopher Hesse, Mark Chen, Eric Sigler, Mateusz Litwin, Scott Gray, Benjamin Chess, Jack Clark, Christopher Berner, Sam McCandlish, Alec Radford, Ilya Sutskever, and Dario Amodei.
\newblock Language models are few-shot learners.
\newblock In \emph{Proc. of Advances in Neural Information Processing Systems (NeurIPS)}, 2020.

\bibitem[Campari et~al.(2020)Campari, Eccher, Serafini, and Ballan]{campari2020exploiting}
Tommaso Campari, Paolo Eccher, Luciano Serafini, and Lamberto Ballan.
\newblock Exploiting scene-specific features for object goal navigation.
\newblock In \emph{Proc. of the European Conference on Computer Vision Workshops (ECCVW)}, 2020.

\bibitem[Campari et~al.(2022)Campari, Lamanna, Traverso, Serafini, and Ballan]{campari2022online}
Tommaso Campari, Leonardo Lamanna, Paolo Traverso, Luciano Serafini, and Lamberto Ballan.
\newblock Online learning of reusable abstract models for object goal navigation.
\newblock In \emph{Proc. of the IEEE/CVF Conference on Computer Vision and Pattern Recognition (CVPR)}, 2022.

\bibitem[Carion et~al.(2020)Carion, Massa, Synnaeve, Usunier, Kirillov, and Zagoruyko]{carion2020end}
Nicolas Carion, Francisco Massa, Gabriel Synnaeve, Nicolas Usunier, Alexander Kirillov, and Sergey Zagoruyko.
\newblock End-to-end object detection with transformers.
\newblock In \emph{European conference on computer vision}, pages 213--229. Springer, 2020.

\bibitem[Chang et~al.(2017)Chang, Dai, Funkhouser, Halber, Niessner, Savva, Song, Zeng, and Zhang]{chang2017matterport}
Angel~X. Chang, Angela Dai, Thomas Funkhouser, Maciej Halber, Matthias Niessner, Manolis Savva, Shuran Song, Andy Zeng, and Yinda Zhang.
\newblock {Matterport3D}: Learning from {RGB-D} data in indoor environments.
\newblock In \emph{Proc. of the International Conference on 3D Vision (3DV)}, 2017.

\bibitem[Chang et~al.(2023)Chang, Gervet, Khanna, Yenamandra, Shah, Min, Shah, Paxton, Gupta, Batra, et~al.]{chang2023goat}
Matthew Chang, Theophile Gervet, Mukul Khanna, Sriram Yenamandra, Dhruv Shah, So~Yeon Min, Kavit Shah, Chris Paxton, Saurabh Gupta, Dhruv Batra, et~al.
\newblock Goat: Go to any thing.
\newblock \emph{arXiv preprint arXiv:2311.06430}, 2023.

\bibitem[Chaplot et~al.(2019)Chaplot, Gandhi, Gupta, Gupta, and Salakhutdinov]{chaplot2020learning}
Devendra~Singh Chaplot, Dhiraj Gandhi, Saurabh Gupta, Abhinav Gupta, and Ruslan Salakhutdinov.
\newblock Learning to explore using active neural {SLAM}.
\newblock In \emph{Proc. of the International Conference on Learning Representations (ICLR)}, 2019.

\bibitem[Chaplot et~al.(2020{\natexlab{a}})Chaplot, Gandhi, Gupta, and Salakhutdinov]{chaplot2020object}
Devendra~Singh Chaplot, Dhiraj~Prakashchand Gandhi, Abhinav Gupta, and Ruslan Salakhutdinov.
\newblock Object goal navigation using goal-oriented semantic exploration.
\newblock In \emph{Proc. of Advances in Neural Information Processing Systems (NeurIPS)}, 2020{\natexlab{a}}.

\bibitem[Chaplot et~al.(2020{\natexlab{b}})Chaplot, Salakhutdinov, Gupta, and Gupta]{chaplot2020neural}
Devendra~Singh Chaplot, Ruslan Salakhutdinov, Abhinav Gupta, and Saurabh Gupta.
\newblock Neural topological slam for visual navigation.
\newblock In \emph{Proc. of the IEEE/CVF Conference on Computer Vision and Pattern Recognition (CVPR)}, 2020{\natexlab{b}}.

\bibitem[Dai et~al.(2017)Dai, Chang, Savva, Halber, Funkhouser, and Niessner]{Dai_2017_CVPR}
Angela Dai, Angel~X. Chang, Manolis Savva, Maciej Halber, Thomas Funkhouser, and Matthias Niessner.
\newblock Scannet: Richly-annotated 3d reconstructions of indoor scenes.
\newblock In \emph{Proceedings of the IEEE Conference on Computer Vision and Pattern Recognition (CVPR)}, 2017.

\bibitem[Das et~al.(2018)Das, Datta, Gkioxari, Lee, Parikh, and Batra]{das2018embodied}
Abhishek Das, Samyak Datta, Georgia Gkioxari, Stefan Lee, Devi Parikh, and Dhruv Batra.
\newblock Embodied question answering.
\newblock In \emph{Proc. of the IEEE/CVF Conference on Computer Vision and Pattern Recognition (CVPR)}, 2018.

\bibitem[Dorbala et~al.(2023)Dorbala, Mullen~Jr, and Manocha]{dorbala2023can}
Vishnu~Sashank Dorbala, James~F Mullen~Jr, and Dinesh Manocha.
\newblock Can an embodied agent find your “cat-shaped mug”? llm-based zero-shot object navigation.
\newblock \emph{IEEE Robotics and Automation Letters}, 2023.

\bibitem[Gervet et~al.(2023)Gervet, Chintala, Batra, Malik, and Chaplot]{modlearn}
Theophile Gervet, Soumith Chintala, Dhruv Batra, Jitendra Malik, and Devendra~Singh Chaplot.
\newblock Navigating to objects in the real world.
\newblock \emph{Science Robotics}, 8\penalty0 (79), 2023.

\bibitem[Gupta and Kembhavi(2023)]{visprog}
Tanmay Gupta and Aniruddha Kembhavi.
\newblock Visual programming: Compositional visual reasoning without training.
\newblock In \emph{Proc. of the IEEE/CVF Conference on Computer Vision and Pattern Recognition (CVPR)}, 2023.

\bibitem[Hu et~al.(2017)Hu, Andreas, Rohrbach, Darrell, and Saenko]{hu2017learning}
Ronghang Hu, Jacob Andreas, Marcus Rohrbach, Trevor Darrell, and Kate Saenko.
\newblock Learning to reason: End-to-end module networks for visual question answering.
\newblock In \emph{Proc. of the IEEE/CVF International Conference on Computer Vision (ICCV)}, 2017.

\bibitem[Hu et~al.(2018)Hu, Andreas, Darrell, and Saenko]{Hu2018StackNMN}
Ronghang Hu, Jacob Andreas, Trevor Darrell, and Kate Saenko.
\newblock Explainable neural computation via stack neural module networks.
\newblock In \emph{Proc. of the European Conference on Computer Vision (ECCV)}, 2018.

\bibitem[Huang et~al.(2022{\natexlab{a}})Huang, Abbeel, Pathak, and Mordatch]{huang2022language}
Wenlong Huang, Pieter Abbeel, Deepak Pathak, and Igor Mordatch.
\newblock Language models as zero-shot planners: Extracting actionable knowledge for embodied agents.
\newblock In \emph{International conference on machine learning}, pages 9118--9147. PMLR, 2022{\natexlab{a}}.

\bibitem[Huang et~al.(2022{\natexlab{b}})Huang, Xia, Xiao, Chan, Liang, Florence, Zeng, Tompson, Mordatch, Chebotar, et~al.]{huang2022inner}
Wenlong Huang, Fei Xia, Ted Xiao, Harris Chan, Jacky Liang, Pete Florence, Andy Zeng, Jonathan Tompson, Igor Mordatch, Yevgen Chebotar, et~al.
\newblock Inner monologue: Embodied reasoning through planning with language models.
\newblock \emph{arXiv preprint arXiv:2207.05608}, 2022{\natexlab{b}}.

\bibitem[Jiang et~al.(2023)Jiang, Sablayrolles, Mensch, Bamford, Chaplot, Casas, Bressand, Lengyel, Lample, Saulnier, et~al.]{jiang2023mistral}
Albert~Q Jiang, Alexandre Sablayrolles, Arthur Mensch, Chris Bamford, Devendra~Singh Chaplot, Diego de~las Casas, Florian Bressand, Gianna Lengyel, Guillaume Lample, Lucile Saulnier, et~al.
\newblock {Mistral 7B}.
\newblock \emph{arXiv preprint arXiv:2310.06825}, 2023.

\bibitem[Johnson et~al.(2017)Johnson, Hariharan, van~der Maaten, Hoffman, Fei-Fei, Zitnick, and Girshick]{Johnson2017InferringAE}
Justin Johnson, Bharath Hariharan, Laurens van~der Maaten, Judy Hoffman, Li Fei-Fei, C.~Lawrence Zitnick, and Ross~B. Girshick.
\newblock Inferring and executing programs for visual reasoning.
\newblock In \emph{Proc. of the IEEE/CVF International Conference on Computer Vision (ICCV)}, 2017.

\bibitem[Khanna et~al.(2024)Khanna, Ramrakhya, Chhablani, Yenamandra, Gervet, Chang, Kira, Chaplot, Batra, and Mottaghi]{khanna2024goat}
Mukul Khanna, Ram Ramrakhya, Gunjan Chhablani, Sriram Yenamandra, Theophile Gervet, Matthew Chang, Zsolt Kira, Devendra~Singh Chaplot, Dhruv Batra, and Roozbeh Mottaghi.
\newblock Goat-bench: A benchmark for multi-modal lifelong navigation.
\newblock In \emph{Proceedings of the IEEE/CVF Conference on Computer Vision and Pattern Recognition}, pages 16373--16383, 2024.

\bibitem[Kolve et~al.(2017)Kolve, Mottaghi, Han, VanderBilt, Weihs, Herrasti, Gordon, Zhu, Gupta, and Farhadi]{kolve2017ai2}
Eric Kolve, Roozbeh Mottaghi, Winson Han, Eli VanderBilt, Luca Weihs, Alvaro Herrasti, Daniel Gordon, Yuke Zhu, Abhinav Gupta, and Ali Farhadi.
\newblock Ai2-thor: An interactive 3d environment for visual ai.
\newblock \emph{arXiv preprint arXiv:1712.05474}, 2017.

\bibitem[Krantz et~al.(2022)Krantz, Lee, Malik, Batra, and Chaplot]{krantz2022instance}
Jacob Krantz, Stefan Lee, Jitendra Malik, Dhruv Batra, and Devendra~Singh Chaplot.
\newblock Instance-specific image goal navigation: Training embodied agents to find object instances.
\newblock \emph{arXiv preprint arXiv:2211.15876}, 2022.

\bibitem[Krantz et~al.(2023)Krantz, Gervet, Yadav, Wang, Paxton, Mottaghi, Batra, Malik, Lee, and Chaplot]{krantz2023navigating}
Jacob Krantz, Theophile Gervet, Karmesh Yadav, Austin Wang, Chris Paxton, Roozbeh Mottaghi, Dhruv Batra, Jitendra Malik, Stefan Lee, and Devendra~Singh Chaplot.
\newblock Navigating to objects specified by images.
\newblock In \emph{Proc. of the IEEE/CVF International Conference on Computer Vision (ICCV)}, 2023.

\bibitem[Li et~al.(2024)Li, Chen, and Lin]{li2024tina}
Dingbang Li, Wenzhou Chen, and Xin Lin.
\newblock Tina: Think, interaction, and action framework for zero-shot vision language navigation.
\newblock \emph{arXiv preprint arXiv:2403.08833}, 2024.

\bibitem[Li et~al.(2023)Li, Li, Savarese, and Hoi]{li2023blip}
Junnan Li, Dongxu Li, Silvio Savarese, and Steven Hoi.
\newblock Blip-2: Bootstrapping language-image pre-training with frozen image encoders and large language models.
\newblock In \emph{International conference on machine learning}, pages 19730--19742. PMLR, 2023.

\bibitem[Liang et~al.(2023)Liang, Huang, Xia, Xu, Hausman, Ichter, Florence, and Zeng]{liang2023code}
Jacky Liang, Wenlong Huang, Fei Xia, Peng Xu, Karol Hausman, Brian Ichter, Pete Florence, and Andy Zeng.
\newblock Code as policies: Language model programs for embodied control.
\newblock In \emph{2023 IEEE International Conference on Robotics and Automation (ICRA)}, pages 9493--9500. IEEE, 2023.

\bibitem[Lin et~al.(2014)Lin, Maire, Belongie, Hays, Perona, Ramanan, Doll{\'a}r, and Zitnick]{lin2014microsoft}
Tsung-Yi Lin, Michael Maire, Serge Belongie, James Hays, Pietro Perona, Deva Ramanan, Piotr Doll{\'a}r, and C~Lawrence Zitnick.
\newblock Microsoft coco: Common objects in context.
\newblock In \emph{Computer Vision--ECCV 2014: 13th European Conference, Zurich, Switzerland, September 6-12, 2014, Proceedings, Part V 13}, pages 740--755. Springer, 2014.

\bibitem[Majumdar et~al.(2024)Majumdar, Ajay, Zhang, Putta, Yenamandra, Henaff, Silwal, Mcvay, Maksymets, Arnaud, et~al.]{majumdar2024openeqa}
Arjun Majumdar, Anurag Ajay, Xiaohan Zhang, Pranav Putta, Sriram Yenamandra, Mikael Henaff, Sneha Silwal, Paul Mcvay, Oleksandr Maksymets, Sergio Arnaud, et~al.
\newblock Openeqa: Embodied question answering in the era of foundation models.
\newblock In \emph{Proceedings of the IEEE/CVF Conference on Computer Vision and Pattern Recognition}, pages 16488--16498, 2024.

\bibitem[Minderer et~al.(2024)Minderer, Gritsenko, and Houlsby]{minderer2024scaling}
Matthias Minderer, Alexey Gritsenko, and Neil Houlsby.
\newblock Scaling open-vocabulary object detection.
\newblock \emph{Advances in Neural Information Processing Systems}, 36, 2024.

\bibitem[Puig et~al.(2023)Puig, Undersander, Szot, Cote, Yang, Partsey, Desai, Clegg, Hlavac, Min, Vondruš, Gervet, Berges, Turner, Maksymets, Kira, Kalakrishnan, Malik, Chaplot, Jain, Batra, Rai, and Mottaghi]{puig2023habitat}
Xavier Puig, Eric Undersander, Andrew Szot, Mikael~Dallaire Cote, Tsung-Yen Yang, Ruslan Partsey, Ruta Desai, Alexander~William Clegg, Michal Hlavac, So~Yeon Min, Vladimír Vondruš, Theophile Gervet, Vincent-Pierre Berges, John~M. Turner, Oleksandr Maksymets, Zsolt Kira, Mrinal Kalakrishnan, Jitendra Malik, Devendra~Singh Chaplot, Unnat Jain, Dhruv Batra, Akshara Rai, and Roozbeh Mottaghi.
\newblock Habitat 3.0: A co-habitat for humans, avatars and robots, 2023.

\bibitem[Radford et~al.(2021)Radford, Kim, Hallacy, Ramesh, Goh, Agarwal, Sastry, Askell, Mishkin, Clark, et~al.]{radford2021learning}
Alec Radford, Jong~Wook Kim, Chris Hallacy, Aditya Ramesh, Gabriel Goh, Sandhini Agarwal, Girish Sastry, Amanda Askell, Pamela Mishkin, Jack Clark, et~al.
\newblock Learning transferable visual models from natural language supervision.
\newblock In \emph{International conference on machine learning}, pages 8748--8763. PMLR, 2021.

\bibitem[Ramakrishnan et~al.(2021)Ramakrishnan, Gokaslan, Wijmans, Maksymets, Clegg, Turner, Undersander, Galuba, Westbury, Chang, et~al.]{ramakrishnan2021habitat}
Santhosh~K. Ramakrishnan, Aaron Gokaslan, Erik Wijmans, Oleksandr Maksymets, Alex Clegg, John Turner, Eric Undersander, Wojciech Galuba, Andrew Westbury, Angel~X. Chang, et~al.
\newblock {Habitat-matterport 3D dataset (HM3D): 1000 large-scale 3D environments for embodied AI}.
\newblock \emph{arXiv preprint arXiv:2109.08238}, 2021.

\bibitem[Ramakrishnan et~al.(2022)Ramakrishnan, Chaplot, Al-Halah, Malik, and Grauman]{ramakrishnan2022poni}
Santhosh~K. Ramakrishnan, Devendra~Singh Chaplot, Ziad Al-Halah, Jitendra Malik, and Kristen Grauman.
\newblock Poni: Potential functions for objectgoal navigation with interaction-free learning.
\newblock In \emph{Proc. of the IEEE/CVF Conference on Computer Vision and Pattern Recognition (CVPR)}, 2022.

\bibitem[Raychaudhuri et~al.(2024)Raychaudhuri, Campari, Jain, Savva, and Chang]{raychaudhuri2024mopa}
Sonia Raychaudhuri, Tommaso Campari, Unnat Jain, Manolis Savva, and Angel~X Chang.
\newblock Mopa: Modular object navigation with pointgoal agents.
\newblock In \emph{Proceedings of the IEEE/CVF Winter Conference on Applications of Computer Vision}, pages 5763--5773, 2024.

\bibitem[Sarlin et~al.(2020)Sarlin, DeTone, Malisiewicz, and Rabinovich]{superglue}
Paul-Edouard Sarlin, Daniel DeTone, Tomasz Malisiewicz, and Andrew Rabinovich.
\newblock Superglue: Learning feature matching with graph neural networks, 2020.

\bibitem[Savva et~al.(2017)Savva, Chang, Dosovitskiy, Funkhouser, and Koltun]{savva2017minos}
Manolis Savva, Angel~X. Chang, Alexey Dosovitskiy, Thomas Funkhouser, and Vladlen Koltun.
\newblock Minos: Multimodal indoor simulator for navigation in complex environments.
\newblock \emph{arXiv preprint arXiv:1712.03931}, 2017.

\bibitem[Savva et~al.(2019)Savva, Kadian, Maksymets, Zhao, Wijmans, Jain, Straub, Liu, Koltun, Malik, et~al.]{savva2019habitat}
Manolis Savva, Abhishek Kadian, Oleksandr Maksymets, Yili Zhao, Erik Wijmans, Bhavana Jain, Julian Straub, Jia Liu, Vladlen Koltun, Jitendra Malik, et~al.
\newblock {Habitat: A platform for embodied AI research}.
\newblock In \emph{Proc. of the IEEE/CVF International Conference on Computer Vision (ICCV)}, 2019.

\bibitem[Shen et~al.(2021)Shen, Xia, Li, Mart{\'\i}n-Mart{\'\i}n, Fan, Wang, P{\'e}rez-D’Arpino, Buch, Srivastava, Tchapmi, et~al.]{shen2021igibson}
Bokui Shen, Fei Xia, Chengshu Li, Roberto Mart{\'\i}n-Mart{\'\i}n, Linxi Fan, Guanzhi Wang, Claudia P{\'e}rez-D’Arpino, Shyamal Buch, Sanjana Srivastava, Lyne Tchapmi, et~al.
\newblock {iGibson 1.0: a simulation environment for interactive tasks in large realistic scenes}.
\newblock In \emph{Proc. of the IEEE/RSJ International Conference on Intelligent Robots and Systems (IROS)}, 2021.

\bibitem[Sur\'is et~al.(2023)Sur\'is, Menon, and Vondrick]{vipergpt}
D\'idac Sur\'is, Sachit Menon, and Carl Vondrick.
\newblock {ViperGPT: Visual Inference via Python Execution for Reasoning}.
\newblock In \emph{Proc. of the IEEE/CVF International Conference on Computer Vision (ICCV)}, 2023.

\bibitem[Team et~al.(2023)Team, Anil, Borgeaud, Wu, Alayrac, Yu, Soricut, Schalkwyk, Dai, Hauth, et~al.]{team2023gemini}
Gemini Team, Rohan Anil, Sebastian Borgeaud, Yonghui Wu, Jean-Baptiste Alayrac, Jiahui Yu, Radu Soricut, Johan Schalkwyk, Andrew~M Dai, Anja Hauth, et~al.
\newblock Gemini: a family of highly capable multimodal models.
\newblock \emph{arXiv preprint arXiv:2312.11805}, 2023.

\bibitem[Touvron et~al.(2023)Touvron, Martin, Stone, Albert, Almahairi, Babaei, Bashlykov, Batra, Bhargava, Bhosale, et~al.]{touvron2023llama}
Hugo Touvron, Louis Martin, Kevin Stone, Peter Albert, Amjad Almahairi, Yasmine Babaei, Nikolay Bashlykov, Soumya Batra, Prajjwal Bhargava, Shruti Bhosale, et~al.
\newblock Llama 2: Open foundation and fine-tuned chat models.
\newblock \emph{arXiv preprint arXiv:2307.09288}, 2023.

\bibitem[Wijmans et~al.(2019)Wijmans, Kadian, Morcos, Lee, Essa, Parikh, Savva, and Batra]{wijmans2019dd}
Erik Wijmans, Abhishek Kadian, Ari Morcos, Stefan Lee, Irfan Essa, Devi Parikh, Manolis Savva, and Dhruv Batra.
\newblock {DD-PPO: Learning near-perfect pointgoal navigators from 2.5 billion frames}.
\newblock In \emph{Proc. of the International Conference on Learning Representations (ICLR)}, 2019.

\bibitem[Williams(1992)]{williams1992reinforce}
Ronald~J Williams.
\newblock Simple statistical gradient-following algorithms for connectionist reinforcement learning.
\newblock \emph{Machine learning}, 8\penalty0 (3):\penalty0 229--256, 1992.

\bibitem[Wu et~al.(2023)Wu, Wang, Xu, Lu, and Yan]{wu2023embodied}
Zhenyu Wu, Ziwei Wang, Xiuwei Xu, Jiwen Lu, and Haibin Yan.
\newblock Embodied task planning with large language models.
\newblock \emph{arXiv preprint arXiv:2307.01848}, 2023.

\bibitem[Ye et~al.(2020)Ye, Batra, Wijmans, and Das]{ye2020auxiliary}
Joel Ye, Dhruv Batra, Erik Wijmans, and Abhishek Das.
\newblock {Auxiliary Tasks Speed Up Learning PointGoal Navigation}.
\newblock In \emph{Proc. of the International Conference on Robot Learning (CoRL)}, 2020.

\bibitem[Ye et~al.(2021)Ye, Batra, Das, and Wijmans]{ye2021auxiliary}
Joel Ye, Dhruv Batra, Abhishek Das, and Erik Wijmans.
\newblock {Auxiliary Tasks and Exploration Enable ObjectNav}.
\newblock In \emph{Proc. of the IEEE/CVF International Conference on Computer Vision (ICCV)}, 2021.

\bibitem[Yokoyama et~al.(2024{\natexlab{a}})Yokoyama, Ha, Batra, Wang, and Bucher]{yokoyama2024vlfm}
Naoki Yokoyama, Sehoon Ha, Dhruv Batra, Jiuguang Wang, and Bernadette Bucher.
\newblock Vlfm: Vision-language frontier maps for zero-shot semantic navigation.
\newblock In \emph{2024 IEEE International Conference on Robotics and Automation (ICRA)}, pages 42--48. IEEE, 2024{\natexlab{a}}.

\bibitem[Yokoyama et~al.(2024{\natexlab{b}})Yokoyama, Ramrakhya, Das, Batra, and Ha]{yokoyama2024hm3d}
Naoki Yokoyama, Ram Ramrakhya, Abhishek Das, Dhruv Batra, and Sehoon Ha.
\newblock Hm3d-ovon: A dataset and benchmark for open-vocabulary object goal navigation.
\newblock \emph{arXiv preprint arXiv:2409.14296}, 2024{\natexlab{b}}.

\bibitem[Yu et~al.()Yu, Kasaei, and Cao]{yul3mvn}
Bangguo Yu, Hamidreza Kasaei, and Ming Cao.
\newblock L3mvn: Leveraging large language models for visual target navigation. in 2023 ieee.
\newblock In \emph{RSJ International Conference on Intelligent Robots and Systems (IROS)}, pages 3554--3560.

\bibitem[Zhou et~al.(2024)Zhou, Hong, and Wu]{zhou2024navgpt}
Gengze Zhou, Yicong Hong, and Qi Wu.
\newblock Navgpt: Explicit reasoning in vision-and-language navigation with large language models.
\newblock In \emph{Proceedings of the AAAI Conference on Artificial Intelligence}, pages 7641--7649, 2024.

\bibitem[Zhou et~al.(2025)Zhou, Hong, Wang, Wang, and Wu]{zhou2025navgpt}
Gengze Zhou, Yicong Hong, Zun Wang, Xin~Eric Wang, and Qi Wu.
\newblock Navgpt-2: Unleashing navigational reasoning capability for large vision-language models.
\newblock In \emph{European Conference on Computer Vision}, pages 260--278. Springer, 2025.

\end{thebibliography}
}

\end{document}